\documentclass[runningheads]{llncs}
\usepackage[T1]{fontenc}
\usepackage{graphicx,verbatim}
\begin{document}
\title{Distribution-based deep multiple instance learning for tumor proportion scoring in NSCLC}
\titlerunning{Distribution-based deep MIL for TPS in NSCLC}
%

\author{Krzysztof Pysz\inst{1}\orcidID{0009-0006-2546-7362} \and
Artur Bartczak\inst{2,3} \and
Jarosław Kwiecień\inst{4} \and 
Piotr Krajewski\inst{4} \and
Witold Dyrka\inst{1}\orcidID{0000-0002-2683-077X}}
\authorrunning{K. Pysz et al.}
%
\institute{Politechnika Wrocławska, Wydział Podstawowych Problemów Techniki, Katedra Inżynierii Biomedycznej, Poland \and
Specjalistyczny Szpital Chorób Płuc w Zakopanem, Zakład Patomorfologii, Poland \and
Narodowy Instytut Onkologii im. Marii Skłodowskiej-Curie, Oddział Kraków, Zakład Patomorfologii Nowotworów, Poland \and
CancerCenter.AI, Wrocław, Poland \\
\email{witold.dyrka@pwr.edu.pl}}
  
\maketitle              
\begin{abstract}
Accurate assessment of tumor proportion score (TPS) in non-small cell lung cancer (NSCLC) is critical for treatment planning and prognosis. Key challenges include the tedious manual work required to annotate each slide, combined with the limited number of experts certified for this task. Multiple instance learning (MIL) has proven to be an effective approach for predicting TPS scores at the slide level; however, existing methods struggle with non-expressive (zero class) images. Our approach involves two models: (1) an embedding-extraction and multiclass-classification network that captures the histopathological features of individual patches, and (2) a MIL model that aggregates these embeddings to predict zero-inflated beta (ZIBeta) parameters representing the overall TPS probability distribution for the entire slide. Using only slide-level TPS scores as labels, we demonstrate how this end-to-end framework can leverage a novel distribution-based architecture to improve prediction accuracy and explainability. ZIBeta modeling significantly outperforms baseline linear and ridge regression while capturing expected accuracy through distribution concentration.


\end{abstract}

\section{Introduction}
Lung cancer, particularly its predominant form, non–small-cell lung carcinoma (NSCLC), is the leading cause of cancer-related death, with most patients diagnosed at an advanced stage and, therefore, ineligible for surgical treatment. In such cases, identifying the histological type of the tumor and evaluating relevant predictive biomarkers enables targeted personalized treatment, which can significantly extend survival compared with standard chemotherapy or radiotherapy.

In advanced NSCLC of any histological type, an essential component of the diagnostic work-up is the assessment of programmed death-ligand 1 (PD-L1) protein expression, as it serves as the therapeutic target of many immune checkpoint inhibitors~\cite{Govindan2022,Sholl2024}. This assessment is performed by a~certified pathologist on immunohistochemistry-stained tissue samples. Evaluating PD-L1 expression is time-consuming and involves determining the tumor proportion score (TPS), defined as the proportion of tumor cells showing positive staining relative to the total number of tumor cells in the histopathological specimen. Because inflammatory cells may also express PD-L1, accurate interpretation requires substantial expertise. Correct TPS quantification has major implications for patient management, with established thresholds often guiding the choice of immunotherapy within the advanced lung cancer treatment program~\cite{ProgramLekowyB6}.

Automated PD-L1 scoring algorithms that largely predict continuous slide-level TPS are the most common and show strong concordance with pathologists~\cite{liu_automated_2021,van_eekelen_comparing_2024}. Alternatively, predictions are often categorized into classes based on clinical thresholds for PD-L1 expression~\cite{hondelink_development_2021,pan_automated_2021,plass2025}, which are determining factors in the future treatment of NSCLC patients. For the three diagnostic intervals, <1, 1--49, and 50--100, the first one is usually the most problematic, with many automated methods experiencing a major decrease in performance when correctly classifying TPS-negative samples~\cite{Kim2024May}. 

The recent emergence of publicly-accessible large IHC datasets for NSCLC~\cite{wang_mihic_2024} enables the employment of weakly-supervised multiple-instance learning (MIL) methods that require minimal labeling.

Additionally, the explainability and uncertainty estimation of any automated machine-learning system are crucial in providing as much contextual information for the pathologist when making the final decision. While attention-based visualization is a common standard in digital pathology, most predictive models lack inherent measures for uncertainty estimation~\cite{Lopez2025preprint}. 

This work addresses these limitations by providing a pipeline for automated TPS prediction using a publicly available multi-stain IHC dataset for tissue classification and embedding extraction~\cite{wang_mihic_2024}, enabling the training of a PD-L1 expression predictor on moderately sized proprietary WSI datasets annotated only with slide-level TPS labels. Additionally, we propose a novel approach to confidence estimation, directly integrated into the model architecture, based on predicting the parameters of a zero-inflated beta (ZIBeta) distribution. 

\section{Materials and methods}

\subsection{Data overview and embedding extraction}
The training dataset for the embedding extraction model was the publicly available MIHIC dataset, containing over 300\,000 images representing 6 tissue types and background images sourced from lung cancer patients~\cite{wang_mihic_2024}. For the TPS prediction model, we used WSIs of IHC-stained histopathological tissue sourced from 162 patients at a lung disease hospital. In 159 cases, the stain used was SP263, with 22C3 being used for the remaining three cases. Each image included a TPS label assigned by a certified pathologist in intervals of 10\%, while more granular data were provided for values under 20\%. 

The reference tissue present in the majority of images from our dataset was removed before the TPS prediction.

For embedding extraction, we trained the \texttt{MobileNetV3Large}~\cite{Howard2019May} architecture on the MIHIC~\cite{wang_mihic_2024} dataset for five epochs using default parameters, except for a~learning rate of $10^{-4}$. Epoch 2 was selected as the one with the lowest validation loss, after which we observed overfitting. Using this architecture, we achieved better overall accuracy than any of the models in the original paper ($0.83$ vs. $0.81$), while matching the top $F1$ score for the \texttt{Tumor} class (0.93). 

\subsection{Multiple-instance learning}
\begin{figure}[!ht]
	\centering
	\includegraphics[width = 0.9\linewidth]{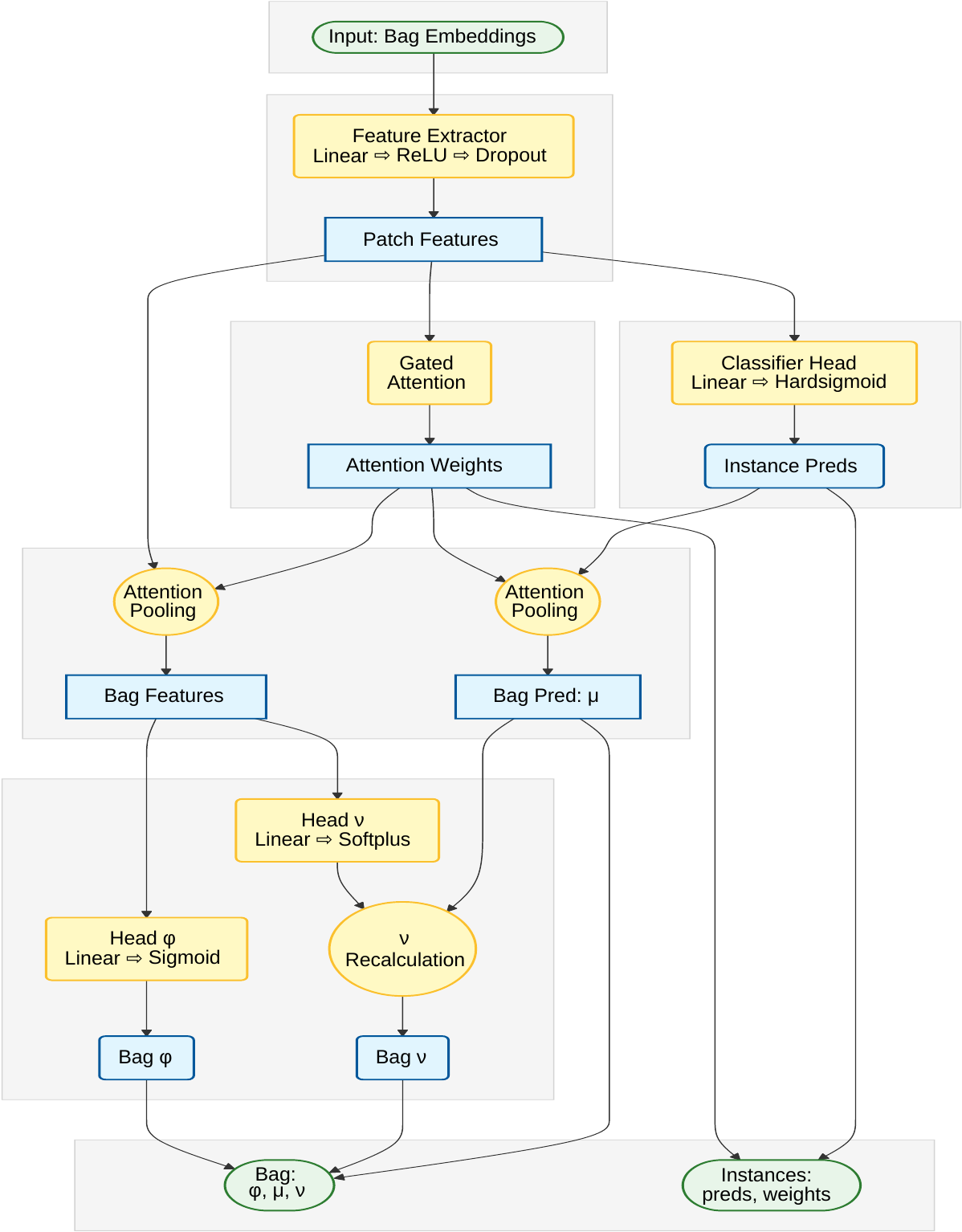}
    \caption{Architecture of ZIBeta model}
    \label{fig:model-arch}
\end{figure}

We propose a MIL architecture consisting of three sequential layers: linear transformation, normalization, and dropout, followed by parallel gated attention-based aggregation~\cite{Ilse2018Feb} and patch-wise TPS prediction (Figure~\ref{fig:model-arch}). The weighted sum of these instance predictions is then used to generate the prediction for the entire bag. To evaluate this approach, we consider the following five models:

\begin{itemize}
\item \textbf{Linear}: \textit{Scikit-learn} \texttt{LinearRegression} using mean pooling over all tumor patch embeddings within a bag.
\item \textbf{Ridge}: \textit{Scikit-learn} \texttt{Ridge} regression using the same mean-pooling approach.
\item $\mathbf{MIL_{base}}$: Basic MIL architecture using mean aggregation of instance predictions instead of attention.
\item $\mathbf{MIL_{att}}$: MIL architecture using gated attention-based aggregation.
\item $\mathbf{MIL_{ZIB}}$: ZIBeta variant extending $\mathrm{MIL_{att}}$ with additional prediction heads. This model takes the bag prediction as the distribution mean $\mu$, and additionally predicts $\phi$ (the probability of TPS=0) and $\nu$ (concentration) of the distribution. The concentration is dynamically recalculated during training to facilitate training stability and guarantee the existence of a~distinct mode:
\[
\nu_\mathrm{corrected} = \nu  + \max\left(\frac{1}{\mu}, \frac{1}{1-\mu}\right) +\epsilon.
\]
\end{itemize}

We used a \texttt{hidden\_dim} of 512 for $\mathrm{MIL_{base}}$ (7 layers, 2.1M parameters), whereas both $\mathrm{MIL_{att}}$ (14 layers, 1.1M parameters) and $\mathrm{MIL_{ZIB}}$ (17 layers, 1.1M parameters) employed a \texttt{hidden\_dim} of 256 with an \texttt{attention\_dim} of 128.

\subsection{ZIBeta model}
Instead of only predicting the actual TPS value, we propose modeling the score as a zero-inflated beta distribution represented by three parameters: $\phi$, which represents the probability of TPS being exactly zero, and $\mu$ and $\nu = (\alpha + \beta)$, which are the mean and concentration of the beta distribution $\mathrm{B}(\alpha, \beta)$. From these, the canonical shape parameters can be calculated as $\alpha = \mu\nu$ and $\beta = (1 - \mu)\nu$, as well as the shifted mean (expected value) $\mu_{\leftarrow}$ and the mode $\omega$:
\begin{eqnarray*}
\mu_{\leftarrow} = (1 - \phi)\mu = (1 - \phi)\,\frac{\alpha}{\alpha+\beta} \\
\omega = \frac{\alpha-1}{\alpha+\beta-2},\quad \alpha > 1,\, \beta > 1.
\end{eqnarray*}
In contrast to predicting the canonical shape parameters $\alpha$ and $\beta$, this alternative parameterization offers more directly interpretable model outputs, with $\mu$ as an approximate predicted value and $\nu$ as an indicator of the certainty of a given prediction. The $\phi$ parameter determines the probability of the outcome being exactly zero, while simultaneously shifting the predicted mean or mode by acting as a damping factor. This enables the ZIBeta model to accurately predict the state of total non-expression of PD-L1, which might be relevant as a diagnostic threshold in some scenarios.

\subsection{Loss functions}
For non-ZIBeta variants, we used the MSE loss, while for the ZIBeta variant, we used a combination of MSE loss and custom negative log-likelihood (NLL) to account for the addition of zero-inflation. For zero-cases, $t = 0$, only $\phi$ was used for loss calculation. In contrast, for TPS-positive cases, $t \in \left(0, 1\right]$, the loss function combined the $\phi$ bias and the probability density function of $\mathrm{B}(\alpha, \beta)$ for the clamped true label~$\tilde{t} = \mathrm{clamp}(t, \epsilon, 1 - \epsilon)$, with $\epsilon=1 \times 10^{-6}$,  as follows:

\begin{eqnarray*}
L_\mathrm{ZIBeta}  & = & -\log L(t) + \mathrm{MSE}((1-\phi)\mu, t),\\
\log L(t) & = & \left\{ 
\begin{array}{ll}
\log(\phi), & \mathrm{if\ } t = 0, \\
\log(1 - \phi) + \log \mathrm{BetaPDF}(\tilde{t}; \alpha, \beta), & \mathrm{if\ } t > 0.
\end{array} 
\right.
\end{eqnarray*}
For the MIL approaches, we introduce entropy minimization on individual patches. This reflects the biological homogeneity of staining within a patch and prevents the models from artificially smoothing predicted scores across all patches:
\[
\Omega = \frac{1}{n} \sum_{i=1}^{n} \mu_{i} (1 - \mu_{i}).
\]
Consequently, the final loss function for both types of models can be expressed as a sum of their native loss function and entropy: $L_\mathrm{final} = L + \Omega$.

\subsection{Training and evaluation protocol}
We trained models on three different train+validation vs. test splits using different seeds. Within each, we randomly shuffled the training and validation splits while keeping the test set identical. The training/validation/test split was 70/15/15, with stratification ensuring similar class distributions. Training was performed for 500 epochs, with a batch size of 32, and the checkpoint with the lowest validation loss was selected as the final model for each configuration. We uniformly used the AdamW optimizer and a dropout rate of 0.3. The learning rate was $5 \times10^{-5}$ for mean-pooling and $2 \times10^{-5}$ for attention-pooling.

We normalized the variable patch counts by randomly downsampling or, less frequently, upsampling to a fixed size of 4096 patches per bag, preserving their spatial coordinates for downstream visualization. Additionally, during training, we randomly sampled 512 patches per bag in each epoch to reduce memory requirements and increase model robustness. Model evaluation was always conducted using all 4096 patches per validation or test slide. 

Model fit was evaluated using Root Mean Square Error (RMSE) and Mean Absolute Error (MAE). Classification performance was assessed via Macro F1, per-class F1 scores ($F1_{0}$, $F1_{1-49}$, and $F1_{50-100}$), and Cohen's Kappa ($\kappa$), which was also used to determine the optimal $\phi$ threshold.

\section{Results}

\subsection{Comparison of the models}
For each model architecture, data split, and run, the epoch resulting in the lowest validation loss was selected for evaluation. The summary metrics were averaged per model architecture, alongside their 95\% CIs. The zero class was defined as TPS expression below 1\%, in accordance with clinical practice.

Due to the binary $\phi$ component in the ZIBeta model, a~decision threshold had to be established. Optimal thresholds were determined for each data split by maximizing Cohen's Kappa on the validation set. The resulting $\phi_\mathrm{val}$ revolved around $0.45 \pm 0.14$ across the splits, with optimal $\kappa_\mathrm{val} = 0.69 \pm 0.09$.

\begingroup
\setlength{\tabcolsep}{3pt}
\renewcommand{\arraystretch}{1.0} 
\begin{table}[!ht]
\caption{Model performance for TPS prediction and classification on the test sets. Results are averaged across all test splits, with 95\% confidence intervals (CIs) shown.}
\centering
\begin{tabular}{|l|c|c|c|c|c|}
\hline
Model & RMSE & MAE & F1$_0$ & F1$_\mathrm{macro}$ & $\kappa$ \\
\hline
Linear & 0.288 $\pm$ 0.013 & 0.207 $\pm$ 0.010 & 0.489 $\pm$ 0.042 & 0.594 $\pm$ 0.020 & 0.442 $\pm$ 0.025 \\
Ridge & 0.265 $\pm$ 0.014 & 0.191 $\pm$ 0.009 & 0.500 $\pm$ 0.050 & 0.586 $\pm$ 0.017 & 0.423 $\pm$ 0.022 \\
\hline
$\mathrm{MIL_{base}}$ & 0.233 $\pm$ 0.024 & 0.165 $\pm$ 0.015 & 0.481 $\pm$ 0.203 & 0.681 $\pm$ 0.054 & 0.618 $\pm$ 0.063 \\
$\mathrm{MIL_{att}}$ & 0.205 $\pm$ 0.032 & 0.136 $\pm$ 0.020 & 0.568 $\pm$ 0.056 & 0.686 $\pm$ 0.032 & 0.582 $\pm$ 0.048 \\
$\mathrm{MIL_{ZIB}}$ & 0.201 $\pm$ 0.029 & 0.140 $\pm$ 0.019 & 0.612 $\pm$ 0.089 & 0.692 $\pm$ 0.045 & 0.573 $\pm$ 0.069 \\
\hline
\end{tabular}
\label{tab:validation-results}
\end{table}
\endgroup

\noindent 
Across all metrics, the ZIBeta model ranked among the best performers compared to alternatives (Table~\ref{tab:validation-results}
). We observed significant improvements in RMSE and MAE for all deep models over the linear and ridge regression baselines. In terms of accuracy on the zero class, both models utilizing gated attention outperformed the regression baselines and the mean-pooling model. However, $\mathrm{MIL_{base}}$ achieved the highest $\kappa$, driven by the superior identification of the 1-49 class.

\subsection{In-depth evaluation of the ZIBeta approach}
\begin{figure}[!ht]
    \centering
    \includegraphics[width=0.85\linewidth]{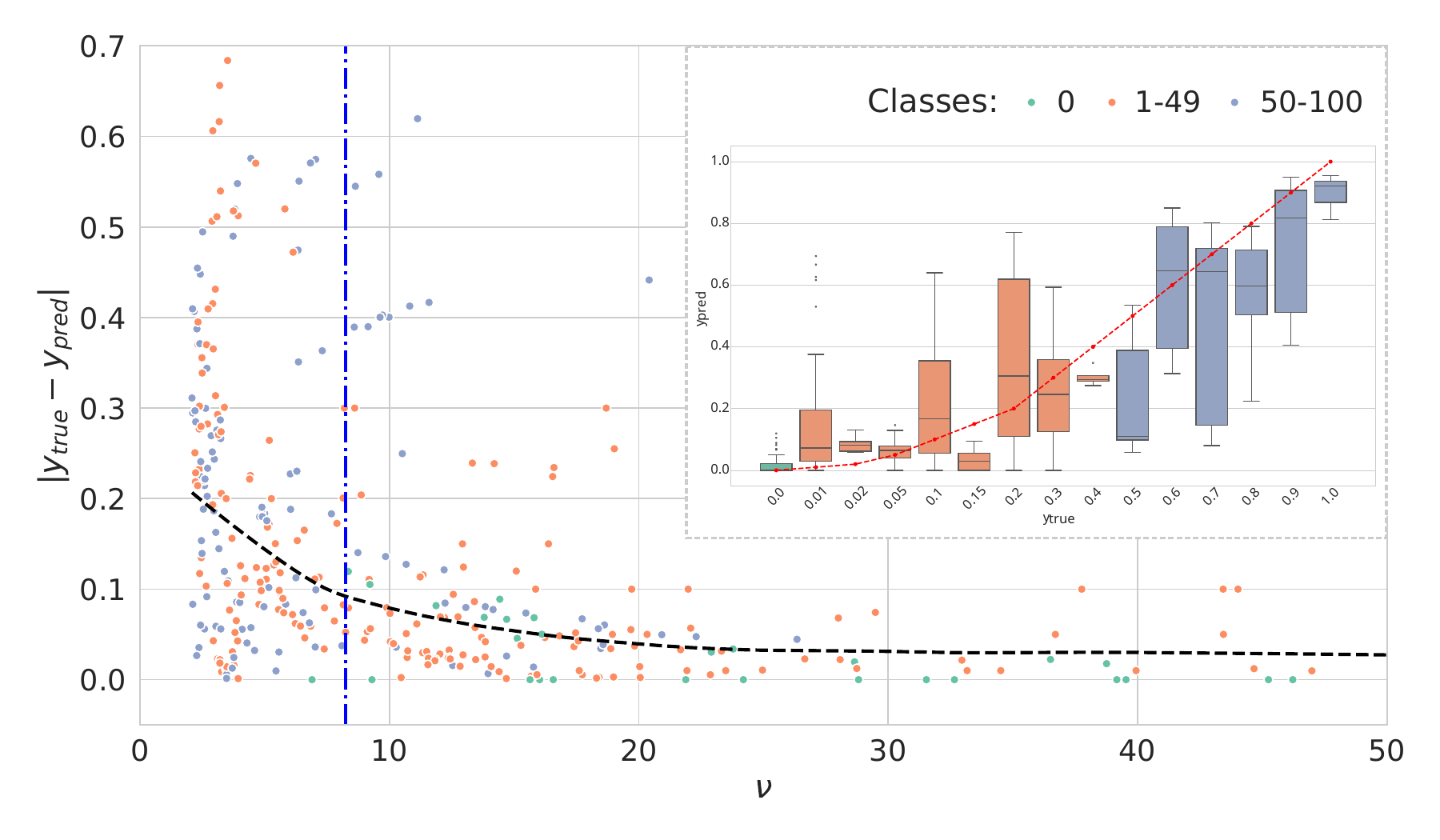}
\caption{LOWESS plots showcasing the relationship between predicted concentration ($\nu$) and absolute error of predictions. Blue line denotes median concentration ($\nu=8.24$) \textbf{Inset:} Distribution of predicted values given the actual label. Red line denotes identity.}
\label{fig:precision-error}
\end{figure}

The theoretical formulation of the ZIBeta model suggests that increasing the precision ($\nu$) should relate to lower prediction errors. To validate this, we plotted the absolute errors of all predictions from the test datasets across all splits for the precision values related to each prediction (Figure~\ref{fig:precision-error}). A significant reduction in absolute errors occurs from a minimum of 2 up to a median of 8.24, with more than a 50\% reduction in estimated errors between these two points. The rate at which the share of each class decreases varies between classes; at the cutoff of median $\nu$, we retain close to 95\% of zero class, 50\% of 1-49 class and only 30\% of 50-100 class. Also prediction range varies significantly between labels, even within a single class, suggesting significant heterogeneity between WSIs and highlighting the need for larger training datasets (Figure~\ref{fig:precision-error} inset).

\begin{figure}[!ht]
\centering

\includegraphics[width=0.85\linewidth]{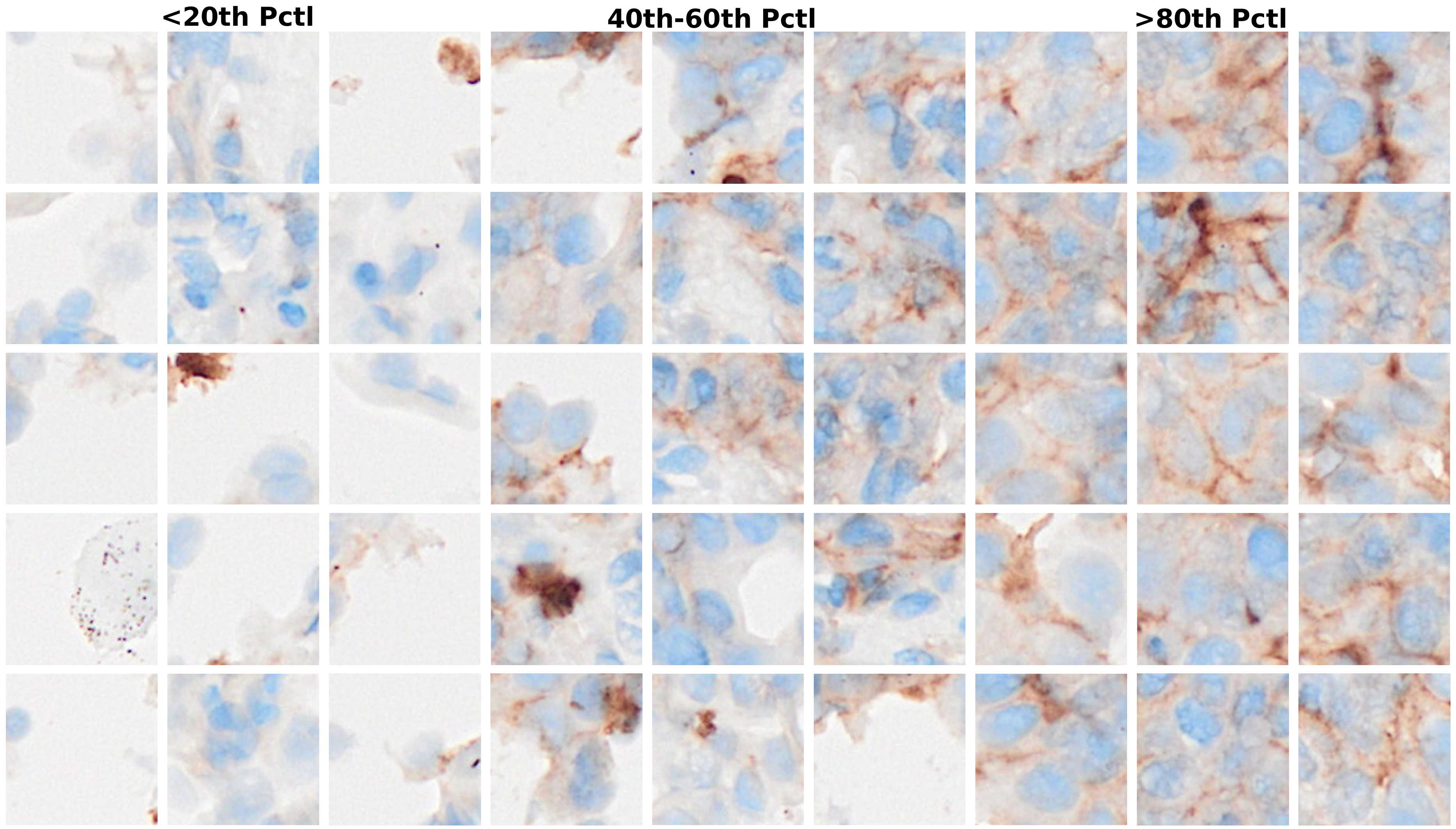}
\caption{Randomly sampled patches belonging to low, medium and high scoring attention categories ordered by percentiles. Each category within corresp. $3\times5$ patch area.}

\label{fig:atts}
\end{figure}

\begin{figure}[!ht]
\centering
\includegraphics[width=0.85\linewidth]{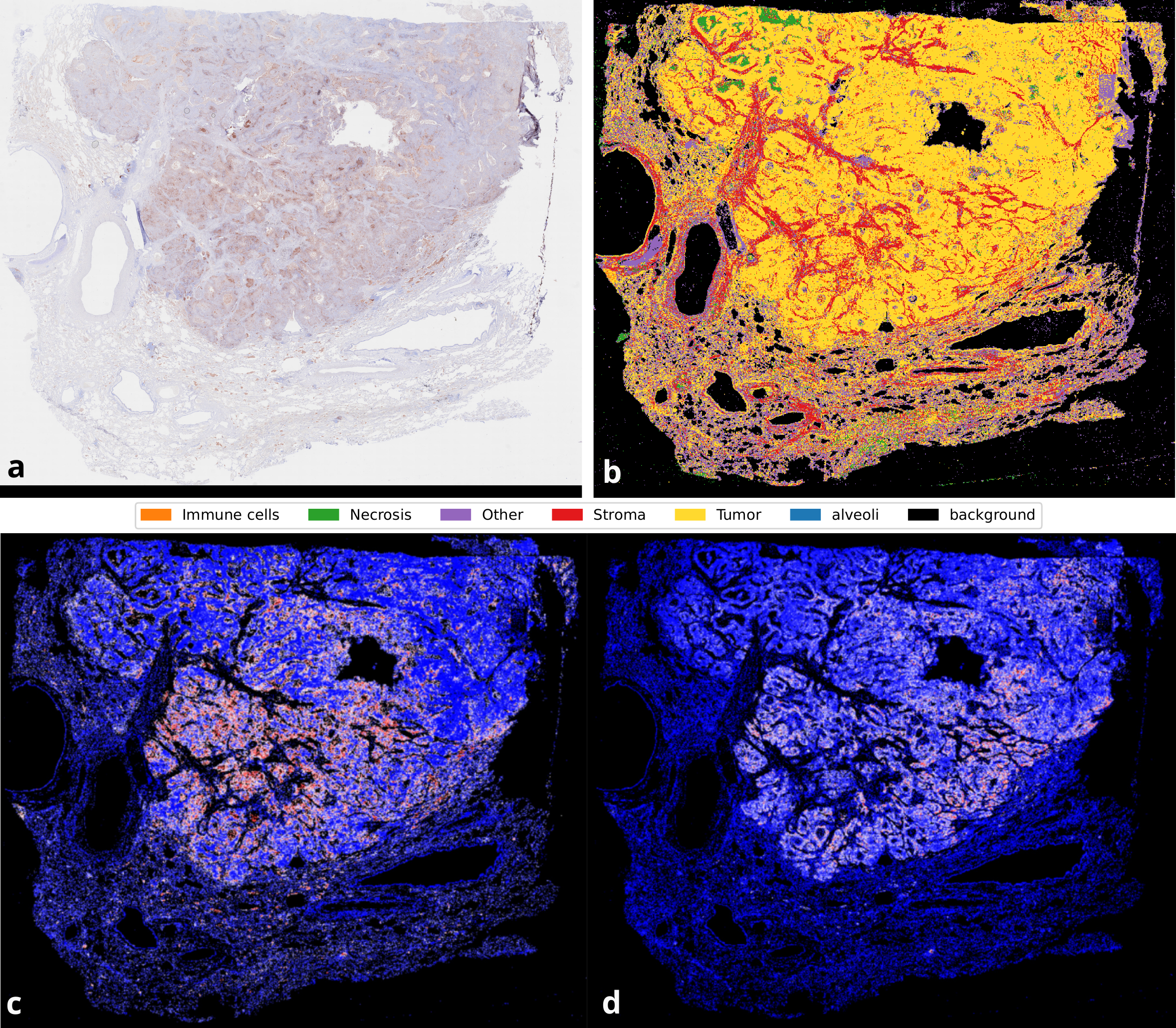}
\caption{Raw image (\textbf{a}) and predicted tissue types (\textbf{b}) for a WSI with TPS label of 60\%. Patch-level TPS (\textbf{c}) and attention (\textbf{d}) extracted with the ZIBeta model by randomly sampling unique 4096-sized batches.}
\label{fig:mihic-atts}
\end{figure}

Additionally, the gated attention mechanism was assessed as a means of filtering out irrelevant patches. We extracted slides for which the true TPS label was relatively close to 50\% to ensure that the slide contained a similar number of stained and non-stained tumor patches. The vast majority of the most attended patches contained stained tumor cells (Figure~\ref{fig:atts}). In general, the attention weight of patches correlates with their information density and, more specifically, with factors such as the ratio of tissue to background, staining prevalence, tumor nuclei, and lack of artifacts. Finally, Figure~\ref{fig:mihic-atts} illustrates the application of the end-to-end framework. Notably, tumor patches at tissue edges, outside of the primary diagnostic region, receive less attention in the final prediction.

\section{Discussion and conclusions}

We designed a weakly supervised TPS prediction pipeline using an open-source dataset for embedding generation, which allows models to train on independent cohorts using only the bag-level TPS label. Our results demonstrate the favorable performance and robustness of ZIBeta modeling for TPS regression in weakly supervised MIL scenarios, especially regarding zero-class accuracy. Distribution-based modeling broadens the scope of uncertainty estimation and prediction visualization. Although the correlation between the ZIBeta concentration parameter and the absolute prediction error is too noisy to serve as a reliable confidence proxy, it highlights the inherent potential of distribution-based approaches over traditional regression. For MIL approaches, the attention mechanism is crucial for noise reduction and guiding the model to focus on information-dense regions. From an explainability perspective, this facilitates identifying disproportionate artifact influence and performing sanity checks on predictions.

Despite these strengths, several limitations warrant further investigation. First, because our dataset is restricted to a single clinical condition (NSCLC) and relies almost exclusively on one staining, it remains unclear how the embeddings used will generalize to other conditions and stains. Second, our true labels were the result of the work of one certified pathologist. This means that we can expect an unknown and consistent bias in the assessment. This uncertainty should be taken into account when analyzing the results, further compounded by the fact that labels for higher TPS are estimated in intervals of 10. 

Although the observed correlation is promising for confidence estimation in MIL regression, the relationship between prediction errors and distribution concentration requires validation on larger, balanced datasets. Further research should investigate strictly non-expressive WSIs as a distinct biological state; the zero-inflation component is uniquely suited for this purpose. Overall, we anticipate the approach will extend to a broad range of weakly supervised tasks.

%
%
%
\bibliographystyle{splncs04}
\bibliography{references}

\end{document}